# *RoboFlamingo-Plus*: Fusion of Depth and RGB Perception with Vision-Language Models for Enhanced Robotic Manipulation

Xiaojian Li[1], Sheng Wang[2], Chao Chen[1], Hailong Wei[1], Yudong Shi[1], Hangjie Mo[1,*]

*Abstract*—As robotic technologies advancing towards more complex multimodal interactions and manipulation tasks, the integration of advanced Vision-Language Models (VLMs) has become a key driver in the field. Despite progress with current methods, challenges persist in fusing depth and RGB information within 3D environments and executing tasks guided by linguistic instructions. In response to these challenges, we have enhanced the existing *RoboFlamingo* framework by introducing *RoboFlamingo-Plus*, which incorporates depth data into VLMs to significantly improve robotic manipulation performance. Our research achieves a nuanced fusion of RGB and depth information by integrating a pre-trained Vision Transformer (ViT) with a resampling technique, closely aligning this combined data with linguistic cues for superior multimodal understanding. The novelty of *RoboFlamingo-Plus* lies in its adaptation of inputs for depth data processing, leveraging a pre-trained resampler for depth feature extraction, and employing cross-attention mechanisms for optimal feature integration. These improvements allow RoboFlamingo-Plus to not only deeply understand 3D environments but also easily perform complex, language-guided tasks in challenging settings. Experimental results show that *RoboFlamingo-Plus* boosts robotic manipulation by 10-20% over current methods, marking a significant advancement.

## I. INTRODUCTION

Recent advancements in Vision Language Models [1]-[3] have marked significant progress, particularly in aligning visual and linguistic information, showcasing an exceptional ability to understand and process complex scenes containing both visual elements and linguistic descriptions. This breakthrough capability of VLMs has garnered widespread attention in the fields of computer vision and natural language processing and sparked interest in applying these advanced models to robotic systems [4]-[6], especially in understanding visual languages. This interest primarily stems from the rich information processing capacity provided by VLMs, enabling robots to interact with humans more naturally and execute vision-language-based tasks with greater accuracy.

Despite the challenges in integrating VLMs as high-level control planners within robotic systems, recent studies, such as RT-1 [4] and RT-2 [5], have demonstrated the immense potential of adapting VLMs for low-level control in robotic systems. These approaches, while effective, come at a high cost. In contrast, the *RoboFlamingo* [7] project offers a cost-effective solution by training on and testing with the Calvin [8] dataset and benchmark, achieving leading performance. However, *RoboFlamingo* still has room for improvement in executing long-term language tasks in unknown environments while adapting to the ABCD-D environment.

Building on the foundation of *RoboFlamingo*, we introduce *RoboFlamingo-Plus*. Below are our contributions to the field:

*1) Multimodal Input Adaptation:* Our model is capable of processing both visual and depth data, enabling a more profound understanding of the environments'3D structure.

*2) Utilizing Pre-trained Resampler Tuning:* We leverage a pre-trained resampler, finely tuned to process depth features as transformed by a vision transformer (ViT), ensuring the effective fusion of depth and RGB features.

*3) Enhanced Feature Integration with Cross-Attention:* Our model employs cross-attention mechanisms to intricately combine linguistic, visual, and depth features, resulting in more precise multimodal processing and enhancing the model's overall performance.

By employing these novel approaches, *RoboFlamingo-Plus* overcomes the input instability associated with fluctuations in RGB information, particularly in unknown environments. The addition of depth data not only strengthens the model's robustness but also supports the execution of complex, long-term language tasks, proving critical for robots operating within dynamic settings. Through the integration of our frozen ViT [9] model and fine-tuned Perceiver Resampler [1] for RGB-D inputs, we set a new benchmark in performance, surpassing current best practices with a performance improvement of 10-20%.

This work not only advances the ability of robotic systems to execute language-guided tasks in complex environments but also opens new avenues for integrating vision language models into practical applications, demonstrating the vast

Authors' information.

potential of multimodal, depth-aware processing in the field of robotics.

## II. RELATED WORK

Language interaction has emerged as an intuitive means of human-robot interaction, enabling non-expert users to control robots to perform multiple tasks through simple commands, thereby avoiding complex programming. This advancement is attributed to the development of pre-trained VLMs, capable of integrating visual information with linguistic instructions, propelling robots' ability to understand and execute language tasks. VLMs have attracted widespread attention in embodied intelligence, signifying a major leap in machine learning's capability to process visual data and natural language. Through pre-training on large-scale dataset, VLMs learn visual features and language patterns, enhancing robots' accuracy and flexibility in complex scenarios.

**From scratch/Fine-tuning:** Initially, Jang et al. [10] and Lynch & Sermanet [11] explored combining vision and language encoders for robotic manipulation. Subsequently, VIMA [12] harnessed the pre-trained T5 [13] model to encode multimodal prompts. Following this, HULC [14] and its enhancement, HULC++ [15], applied fine-tuning to vision and language encoders on the CALVIN dataset. Brohan and others (2022) introduced RT-1 [4], a 35M parameter robotic transformer model integrating vision, language, and action (VLA). RT-1 leveraged a vast real-world manipulation dataset, tokenizing actions and aligning them with vision and language using the Universal Sentence Encoder [16] and EfficientNet-B3 [17] for language embedding and visual tokenization. Meanwhile, The MT-ACT [18] employed a Transformer [19] and CVAE [20] with FiLM-based [3] conditioning for precise multimodal data interpretation, achieving task generalization and a 40% performance improvement in new environments.

**Co-fine-tuning:** In the evolution of co-fine-tuning methodologies, Driess [21] showcased the 540B PaLM-E model, a groundbreaking utilization of pre-trained VLM that blends web-sourced visual language data with robotic manipulation datasets for an all-encompassing training approach. Building upon this, Brohan introduced RT-2 [5], leveraging substantial vision language architectures like PaLI-X [22] and PaLM-E, to conduct training that combines internet data with specific robotic manipulation scenarios, underscoring the adaptability of VLMs in robotic tasks. Shifting away from purely web-based co-fine-tuning, RoboPlan [23] introduced a dataset focusing on long-horizon planning, incorporating data from both robotic and human actions to enhance model accuracy and data gathering efficiency for complex sequences. In tandem, Zipeng Fu et al.'s Mobile ALOHA [6] project demonstrated the value of co-training with assorted robotic datasets on mobile manipulation tasks, utilizing a sophisticated whole-body teleoperation system to significantly improve success rates, thereby highlighting the advantages of utilizing diverse robotic datasets in the co-fine-tuning of VLMs for intricate manipulation challenges. However, the reliance of these strategies on extensive, often proprietary datasets introduces limitations to their scalability and general applicability.

To address the limitations posed by the reliance on extensive, *RoboFlamingo* [7] leverages OpenFlamingo's [2] pre-trained parameters, sourced from extensive web multimodal data, and fine-tunes them with robotic-specific datasets. This process includes enhancements in critical areas such as cross-attention, Perceiver Resampler, and Long Short-Term Memory (LSTM) [24], propelling *RoboFlamingo* to the forefront of robotic manipulation technologies. While many existing robot models primarily rely on RGB data to predict robot actions, the potential of depth information has not been fully exploited in the application of existing large models. We believe that appropriately introducing depth information can significantly enhance the model's robustness and task execution capability in unknown environments.

Building on this, our focus shifts to how depth information can be effectively integrated into existing models to address the current methods' shortcomings in applying depth information. Through carefully designed improvements, we aim not only to enhance the model's three-dimensional perception of the environment but also to expand the robot model's application range in more complex scenarios. This improvement intends to deepen the understanding of environmental dimensions and, by introducing depth information, further enhance *RoboFlamingo*'s accuracy and flexibility in robot action prediction and task execution.

## III. ARCHITECTURE

Imitation [25] learning enables an agent to replicate manipulation strategies from expert data labeled with instructions, represented as $\mathcal{D} = \{(\tau, l)\}_{i=0}^{D}$, where $D$ denotes the count of trajectories, $l$ is the linguistic instruction, and $\tau = \{(o_t, a_t)\}$ consists of prior observed states and actions necessary to achieve the goal outlined by the instruction. The learning goal is succinctly defined as achieving a maximum likelihood objective for goal-conditioned imitation to derive the policy $\pi_\theta$:

$$\ell = \mathbb{E}_{(\tau,l)_i \sim \mathcal{D}}[\sum_{t=0}^{|\tau|} log \pi_\theta(a_t|o_t, l)]$$

In traditional settings, the input space $o_t$ primarily includes RGB data, which has significant potential for improvement in complex real-world manipulation tasks. To optimize adaptability and robustness, we fuse depth and RGB information into *RoboFlamingo-Plus*, allowing the model to more accurately interpret spatial relationships and physical interactions.

Fig. 1 showcases the advanced framework of *RoboFlamingo-Plus*, highlighting its architecture and enhanced input modalities. This visual demonstrates how the addition of depth data complements RGB inputs, significantly boosting the model's performance.

### A. Language-conditioned Robot Control

The problem of language-conditioned robot control can be modeled as a goal-conditioned partially observable Markov decision process [26]: $\mathcal{M} = \langle \mathcal{S}, \mathcal{O}, \mathcal{A}, \mathcal{T}, \rho_0, \mathcal{L}, \emptyset, f \rangle$, where $\mathcal{S}$ and $\mathcal{O}$ are the set of states and observations separately, $\mathcal{A}$ is the action space, $\mathcal{T}: \mathcal{S} \times \mathcal{A} \to \mathcal{S}$ is the environment dynamics function, $\rho_0: S \to [0, 1]$ is the initial state distribution, $\emptyset(s)$ indicate if the task is successful, and $f(o|s): S \to O$ is the observation function. Specifically, for each controlling episode, the robot is given a goal, represented by a length-M free-form language instruction $l \in \mathcal{L}$ at every timestep $t$. The observations $o_t$ are typically two images $I_t^{rgb}$, $G_t^{rgb}$ and two depth $I_t^{depth}$, $G_t^{depth}$, sourced from a third-perspective camera and a camera mounted on the robot's end-effector, respectively. The controlling policy can be modeled as a goal-conditioned policy $\pi(a|o, l): S \times \mathcal{L} \to \mathcal{A}$, and the action $a$ is typically the desired relative position and pose of the gripper that mounted on the robot's end-effector, along with the gripper's open/close status.

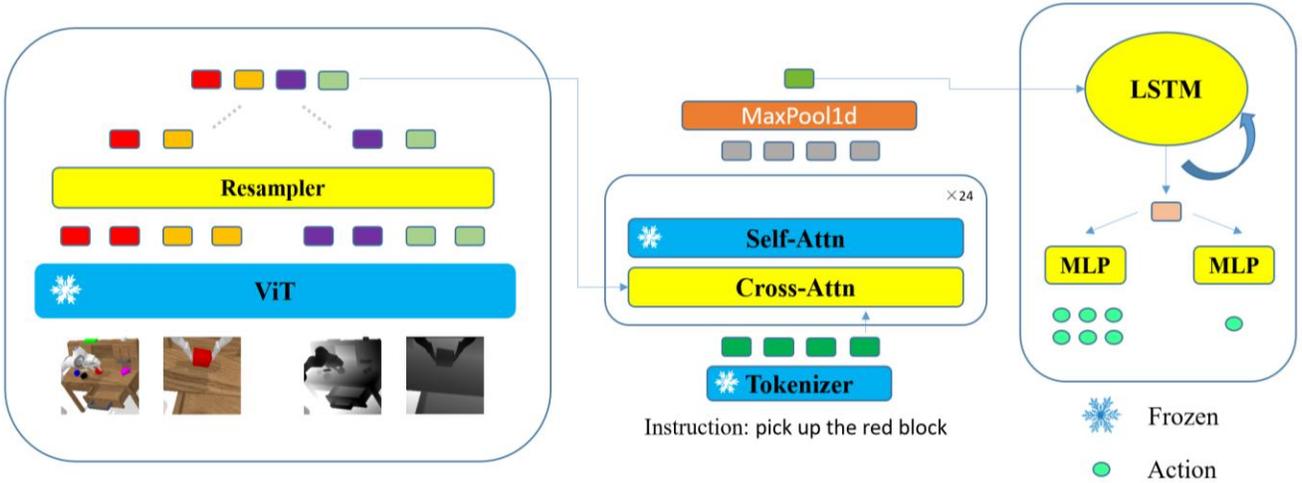

Fig. 1 The schematic diagram of the proposed *RoboFlamingo-Plus* starts with encoding two RGB and two processed depth images using a pretrained ViT. The encoded outputs pass through a pretrained, fine-tunable Perceiver resampler. Sequences from each group are concatenated and input into the key-value (KV) fully connected layers of the cross-attention mechanism, designed for multimodal interactions. Language instructions are tokenized and encoded, serving as the query (Q) in cross-attention, interacting with multimodal K, V pairs formed from concatenated RGB and depth sequences. This interaction facilitates attention-based processing between language and visual modalities. The cross-attention sequence is further refined through a self-attention mechanism with fully connected layers for Q, K and V. This combination of cross- and self-attention is repeated 24 times, each cycle enhancing multimodal data integration. After 24 attention cycles, a MaxPool1d layer filters the sequence to select the most relevant features. The filtered sequence is fine-tuned through four LSTM layers to enrich temporal dynamics. Finally, the sequence is translated into robotic commands via two sets of MLPs, converting the processed multimodal data into actions of a robotic arm as per the input instructions.

In our *RoboFlamingo-Plus*, the policy $\pi_\theta(a|o, l)$ is parameterized by $\theta$. It consists of a backbone based on Flamingo $f_\theta$ and a policy head $p_\theta$. The backbone takes visual observations and language-represented goals as the input and provides a latent fused representation at each timestep for the policy head: $X_t = f_\theta(o_t, l)$. Then the policy head further predicts the action to fulfill the specified goal for the robot: $a_t = p_\theta(X_t, h_{t-1})$, where $h_{t-1}$ is the hidden state from the last step that encodes the history information for decision-making. We will introduce each module in detail in the following sections.

### B. Depth Fusion Flamingo Backbone

The *RoboFlamingo-Plus* backbone $f_\theta$ is utilized to process vision and depth inputs along with language inputs at each decision-making stage. Visual and depth data are transformed into a latent state by the vision and depth encoder, which is subsequently integrated with linguistic objectives using the feature fusion decoder, initiated with a pre-trained language model.

#### 1) Vision And Depth Encoder

**Vision encoder:** The vision encoder incorporates a ViT. At each timestep $t$, the visual observations $I_t^{rgb}$ and $G_t^{rgb}$ are transformed into a sequence of visual tokens $\hat{X}_t^v$ via the ViT:

$$\hat{X}_t^v = \text{ViT}(I_t^{rgb}, G_t^{rgb}),$$

where $\hat{X}_t^v = (\hat{X}_{t1}^v, \cdots, \hat{X}_{tN}^v)$ denotes the sequence of visual tokens at time $t$, and $N$ is the number of tokens in the encoded output.

**Depth encoder:** In our enhanced framework, *RoboFlamingo-Plus*, for each depth map $D$, we first replicate it three times to obtain $D_1, D_2, D_3$, where $D_1 = D_2 = D_3 = D$. Subsequently, we normalize each pixel value $d_{ij}$ in every depth map by calculating the maximum $D_{max}$ and minimum $D_{min}$ depth values identified across default dataset, resulting in the normalized depth value $d'_{ij}$.

$$d'_{ij} = \frac{d_{ij} - D_{min}}{D_{max} - D_{min}},$$

Following the normalization process, we conduct a further step to standardize each depth map for *RoboFlamingo-Plus*. This step involves calculating the mean $\mu$ and standard deviation $\sigma$ of the normalized depth values across the default dataset, providing a basis for the regularization process. Specifically, for each pixel value $d'_{ij}$ in the normalized depth map, the regularization process is applied as follows:

$$d''_{ij} = \frac{d'_{ij} - \mu}{\sigma},$$

Here, $d''_{ij}$ represents the regularized depth value, which has been standardized to have consistent mean and variance,

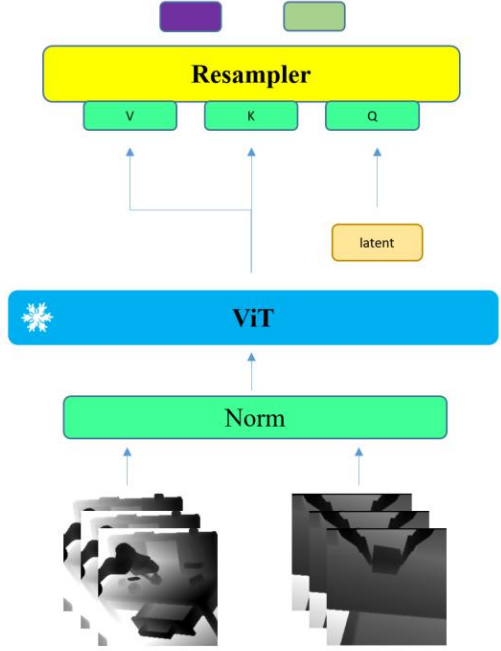

Fig. 2 This is processing flowchart of the depth map we proposed. Firstly, make three copies of the depth map, normalize it, and extract its features using frozen VIT-L-14. Then, use a pre-trained resampler to fine tune the features obtained from VIT.

enhancing the model's ability to interpret depth information uniformly.

As shown in Fig. 2, This regularization process ensures that the depth information fed into *RoboFlamingo-Plus* is not only normalized with respect to global depth extremities but also standardized across the default dataset, facilitating a more reliable and uniform interpretation of depth cues by the model. Through these statistical normalizations and regularizations, we prepare the depth images for effective encoding by the ViT, enabling *RoboFlamingo-Plus* to leverage depth information alongside RGB data for improved environmental interaction and manipulation task performance.

For the depth image processing, the regularized depth images $d''_{ij}$, specifically those processed as $I_t^{depth}$ and $G_t^{depth}$, are prepared for encoding by the ViT, undergoing a unique preparation process. Initially, each depth image is replicated three times, and each copy is normalized against the maximum and minimum values identified across the default dataset. This normalization step adjusts the depth images based on a comprehensive understanding of depth variations within the dataset. Subsequently, each depth image, now regularized with consistent mean and variance, is fed into a pre-trained ViT, akin to the RGB images, to generate a depth-encoded output. This methodical approach ensures that the depth information, now standardized and ready for intricate processing, is optimally utilized by *RoboFlamingo-Plus* for enhanced task execution and environmental navigation.

$$\hat{X}_t^{de} = \text{ViT}(I_t^{depth},\ G_t^{depth}),$$

where $\hat{X}_t^{de} = (\hat{X}_{t1}^{de}, \cdots, \hat{X}_{tN}^{de})$ denotes the sequence of depth tokens at time $t$, and $N$ is the number of tokens in the encoded output.

*2) Perceiver Resampler*

Due to the findings presented in the Flamingo [1] studies, it has been demonstrated that freezing parameters of the ViT contributes to stability in performance, while training the ViT further could lead to catastrophic forgetting. Therefore, our model similarly opts to freeze the ViT. However, the pre-trained parameters of the ViT, developed on web datasets, lack the inherent capability to extract features specific to robotic arms' RGB and depth information. Consequently, there is a necessity for a component to fine-tune the features output by the ViT. The Perceiver Resampler, utilizing an attention mechanism and a set of learnable parameters, is well-suited for this task. Moreover, the resampler's dimensionality reduction capability, compressing the sequence length from $N$ to $K$, not only minimizes the data required for training but also effectively learns and retains crucial feature information. Formally, the RGB and depth partial resampling process are described as:

$$K_R^v = \hat{X}_t^v W_K^R, V_R^v = \hat{X}_t^v W_V^R, X_t^v = softmax(\frac{Q_R K_R^{v^T}}{\sqrt{d}})V_R^v,$$

$$K_R^{de} = \hat{X}_t^{de} W_K^R, V_R^{de} = \hat{X}_t^{de} W_V^R, X_t^{de} = softmax(\frac{Q_R K_R^{de^T}}{\sqrt{d}})V_R^{de},$$

with $Q_R \in \mathbb{R}^{K \times d}$ acting as the query vector corresponding to the resampler's learnable parameters, d being the size of the hidden dimension, and $W_K^R, W_V^R \in \mathbb{R}^{d_{v/de} \times d}$ representing the linear transformation matrices for key and value, respectively, where $d_{v/de}$ is the dimension of the visual or depth token features. Consequently, $K_R^v$ and $V_R^v$, along with $K_R^{de}$ and $V_R^{de}$, emerge as the transformed key and value vectors tailored for processing vision and depth inputs, respectively.

The final step in our enhanced coding process is to merge the RGB and depth coding after VIT and resampling to create a unified visual and depth feature set. This merged feature set will then be utilized by the robotic flamingo model for subsequent decision making and action generation to navigate and manipulate the environment more efficiently using a comprehensive understanding of visual and depth information.

$$X_t^{v-de} = concat(X_t^v, X_t^{de}),$$

*3) Feature Fusion Decoder*

The resampler outputs compressed visual and depth tokens, $X_t^{v-de} \in \mathbb{R}^{K \times d}$, which are then input into the feature fusion decoder. This decoder is tasked with creating a vision-language joint embedding by integrating the language instruction with the encoded vision and depth features, $X_t^{v-de}$. In *RoboFlamingo-Plus*, the decoder, borrowed from OpenFlamingo, is adapted according to the methodology described by Awadalla et al [2]. The decoder is structured with $L$ layers, incorporating a transformer decoder layer and a cross-attention layer in each. The transformer layers, sourced from pre-trained language models like MPT [27], remain unchanged throughout training. The cross-attention layer, which treats language tokens as queries and visual tokens as keys and values, undergoes fine-tuning based on imitation learning objectives using manipulation data (refer to subsequent sections for details). Denoting $x_i \in \mathbb{R}^d$ as the word embedding token for the *i*-th instruction token, with $M$ being the instruction length and $X \in R^{M \times d}$ as the instruction's embedded matrix, in which the embedded natural language instruction is $X = (x_1, x_2, \cdots, x_M)$. The output $X_t^{l+1}$ from the *l*-th decoder layer, given input $X_t^l$, is calculated as:

$$\hat{X}_t^l = Tanh(\alpha) \cdot MLP\left(A(X_t^l W_Q^C, X_t^{v-de} W_K^C, X_t^{v-de} W_V^C)\right) + X_t^l,$$

$$X_t^{l+1} = MLP\left(A(\hat{X}_t^l W_Q^S, \hat{X}_t^l W_K^S, \hat{X}_t^l W_V^S)\right) + \hat{X}_t^l,$$

where $X_t^l = X$, $\hat{X}_t^l$ denotes the gated cross-attention layer's output at $t$, $W_Q^C, W_K^C, W_V^C \in \mathbb{R}^{d \times d}$ are the learnable parameters for the cross-attention layer. $\alpha \in \mathbb{R}$ is a learnable gate parameter adjusting the mix for stability, $W_Q^S, W_K^S, W_V^S \in \mathbb{R}^{d \times d}$ are the self-attention layer's learnable parameters, and Multilayer Perceptron (MLP) signifies a multi-layer perceptron network. Through deep integration of vision and language tokens, the final output $X_t = X_t^L = \{X_{t,1}^L, X_{t,2}^L, \ldots, X_{t,M}^L\}$ at timestep *t* aims to provide a rich vision-language joint embedding for robot manipulation tasks.

Table 1. Task 1-5 reflects the success rates of executing specific tasks. It displays a performance comparison where our *RoboFlamingo-Plus* model outperforms the original, achieving a performance improvement of 10-20%. Success rates were evaluated across 200 sequences from the Calvin dataset for each configuration. The label 'Enriched' denotes tests conducted with a variety of linguistic inputs to assess the model's robustness to language variations.

| Model | Train | Test | Task1 | Task2 | Task3 | Task4 | Task5 | Avg |
|---|---|---|---|---|---|---|---|---|
| *RoboFlamingo* | ABCD | D | 0.96 | 0.87 | 0.78 | 0.705 | 0.625 | 3.94 |
| Ours | ABCD | D | **0.965** | **0.875** | **0.805** | **0.75** | **0.655** | **4.02** |
| *RoboFlamingo* | ABC | D | 0.74 | 0.57 | 0.37 | 0.285 | 0.205 | 2.062 |
| Ours | ABC | D | **0.90** | **0.73** | **0.575** | **0.46** | **0.315** | **3.00** |
| *RoboFlamingo* | ABC | D(Enriched) | 0.46 | 0.205 | 0.095 | 0.055 | 0.015 | 0.83 |
| Ours | ABC | D(Enriched) | **0.605** | **0.39** | **0.255** | **0.15** | **0.07** | **1.42** |

### C. Policy Head

The output $X_t^L$ from the feature fusion decoder is developed as the combined representation of both vision observation and language instruction, which is then converted into low-level control commands. To facilitate this conversion, we employ an additional policy head $p_\theta$ for action prediction, such as the 7 degrees of freedom (DoF) for the end-effector pose and the state of the gripper. We explore various methods to process historical observation sequences for the policy head, including using a LSTM [24] network coupled with a (MLP) for the final action prediction, a decoder-only Transformer with an MLP, or a standalone MLP focusing on single-step data (detailed further in Section 5). For instance, in the LSTM approach, with the joint vision-language embedding sequence $X_t^L$, we apply a max-pooling operation across the token dimension to derive an aggregated embedding, from which the action is predicted as follows:

$$\hat{X}_t = MaxPooling(X_t),$$

$$h_t = LSTM(\hat{X}_t, h_{t-1}); a_t^{pose}, a_t^{gripper} = MLP(h_t),$$

where $h_t$ is the hidden state at time t, and $a_t^{pose}$, $a_t^{gripper}$ are the predicted pose of the end-effector and the gripper's state, respectively.

### D. Training Objective

To refine the pre-trained backbone and policy head, we apply maximum likelihood objectives for imitation learning. Specifically, we target optimizing the relative pose through regression loss (employing mean squared error (MSE) loss) and classify the gripper's state using classification loss (utilizing binary cross-entropy (BCE) loss):

$$\ell = \sum_t MSE(a_t^{pose}, \hat{a}_t^{pose}) + \lambda_{gripper} BCE(a_t^{gripper}, \hat{a}_t^{gripper}),$$

where $\hat{a}_t^{pose}$, $\hat{a}_t^{gripper}$ represent the demonstrated end-effector pose and gripper status at time *t*, respectively, and $\lambda_{gripper}$ is the weight of the gripper loss.

During training, we adhere to the fine-tuning approach of OpenFlamingo, focusing on adjusting the parameters of the resampler, each decoder layer's gated cross-attention module, and the policy head, while keeping the rest of the parameters fixed.

## IV. EXPERIMENTS

In this research, we investigate robotic manipulation tasks where the robot, lacking exact environmental state awareness, leverages visual and depth information, and proprioception from multiple camera angles. Our experiments employ CALVIN, an open-source benchmark for learning tasks conditioned on long-horizon language instructions, focusing on a Franka Emika Panda robotic arm with 7 degrees of freedom to navigate through the tasks. CALVIN presents a comprehensive testbed with 34 diverse tasks and over 1000 instructional sequences for sequential task execution. The dataset, divided into four environmental splits, includes over 2 million trajectories from 6 hours of human-operated sessions, with only 1% annotated with language instructions. Our experimental setup utilized eight A-800 GPUs for training spanning 3-5 days and conducted tests on 200 sequences to validate the reliability of our results, showcasing the policy's effectiveness in sequential goal-driven tasks and its robustness in handling the dataset's inherent suboptimal behaviors.

In the CALVIN benchmark, we compare RoboFlamingo-Plus with the established RoboFlamingo, which utilizes OpenFlamingo's pre-trained model, augmented

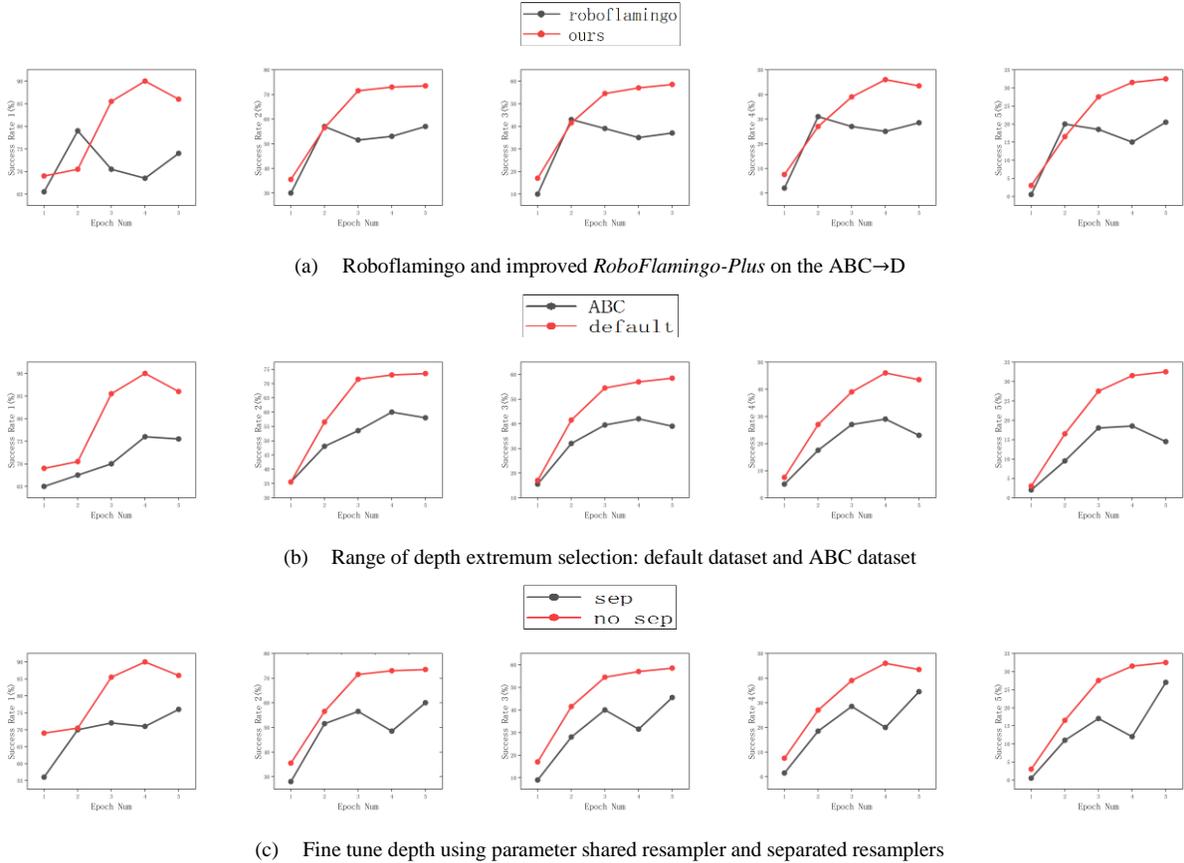

(a) Roboflamingo and improved *RoboFlamingo-Plus* on the ABC→D

(b) Range of depth extremum selection: default dataset and ABC dataset

(c) Fine tune depth using parameter shared resampler and separated resamplers

Fig. 3 From left to right are the corresponding success rates of the model at each epoch corresponding to successive execution of tasks 1 to 5, respectively. (a) It is a comparison between our model and the original model to demonstrate the effectiveness of our model improvement. (b) This is our comparative experiment on deep extreme value processing, which proves that the selection of deep extreme values can affect the performance of the model to a certain extent, with a moderate range being better. (c) This is a ablation experiment we conducted on a resampler to test its fine-tuning ability. We found that a resampler can simultaneously perform tasks of fine-tuning RGB and depth.

by an external LSTM policy head. This baseline predicts actions based on two main inputs: RGB data and language instructions.

Our experimental evaluation proceeds in three phases to verify RoboFlamingo-Plus's effectiveness. First, we assess the model's imitation abilities, demonstrating its aptitude for following language-directed tasks. The second phase examines the model's zero-shot generalization, highlighting its adaptability to tasks and contexts not encountered during training. The final phase involves ablation studies that carefully evaluate each architectural element's importance, underscoring their critical roles in achieving the model's top-tier performance.

### A. Imitation Performance

*RoboFlamingo-Plus* shows slight improvements over its predecessor when trained and tested in the ABCD to D environment, indicating that the RGB modality is already well-optimized for imitation learning tasks in this dataset. Importantly, adding depth features to RoboFlamingo-Plus does not compromise its performance, as seen in Table 1. This enhancement confirms that depth sensing strengthens the model's robustness in familiar environments without diminishing its existing imitation learning abilities.

### B. Zero-Shot Generalization

To assess the zero-shot generalization ability, we evaluate *RoboFlamingo* in two aspects: vision-depth and language.

For vision-depth generalization, we train models on splits A, B, and C and test on split D, which presents a different vision context. Our method significantly outperforms baselines in this vision generalization scenario (ABC → D), as shown in Table 1.

In the context of language generalization, we enhance the linguistic framework by creating 50 synonymous instructions for each task using GPT-4 [28], from which instructions are randomly sampled during evaluation. Our approach maintains a performance lead of 10-20% over *RoboFlamingo* in this enhanced language generalization setting.

### C. Ablation Studies

*1) Sep_resampler*

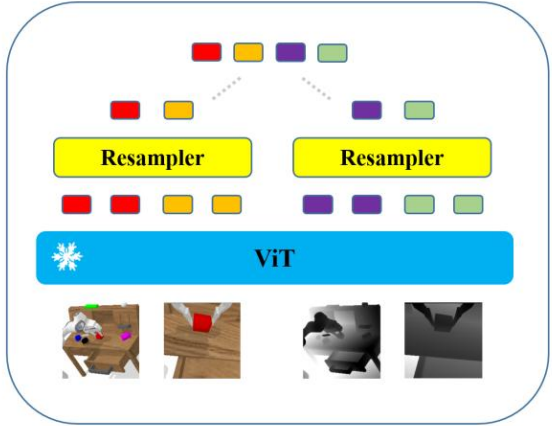

Fig. 4: A resampler can already achieve good results. We want to test not sharing parameters and train two resamplers simultaneously, one for fine-tuning RGB features and the other for fine-tuning depth features.

To explore the efficacy of using separate for distinct modality data, we initially observed that a single resampler could adequately process combined RGB and depth information. Encouraged by these results, we then investigated whether separate resampling of RGB and depth modalities would enhance performance. In this subsequent study, models were initialized with identical pre-trained parameters for two resamplers, which were then fine-tuned independently without parameter sharing for the specialized processing of RGB and depth features. Fig. 4 presents the schematic of this separate resampler framework, highlighting the dedicated handling of each modality. The task success rates achieved through this methodological variation are detailed in Fig. 3, offering insight into the comparative benefits of distinct resampling strategies.

*2) Depth Extremes*

In our experiments, we observed that varying maximum and minimum depth values significantly impacts task execution success rates. We hypothesize that a large gap between these values may reduce differentiation after normalization, making it difficult for the model to discern precise depth information, thus lowering success rates. To investigate, we tested both a default dataset with fewer operations and a narrower range of depth extremes compared to the comprehensive ABC dataset, and the full ABC dataset to determine their respective maximum and minimum depth values. The default dataset's narrower depth range makes it more sensitive to variations, offering insight into the effect of depth value selection on model performance.

We evaluated success rates after normalizing depth values using the mean and variance from both the default and ABC datasets. The results are shown in Fig. 3, and Fig. 5 depicts the normalization's effect on depth perception sensitivity, providing context on how the selection of depth range affects task success.

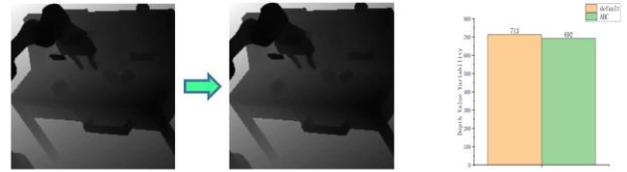

(a) Depth Sensitivity Analysis

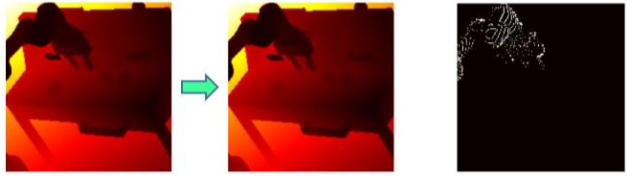

(b) Default dataset: Extremely small range

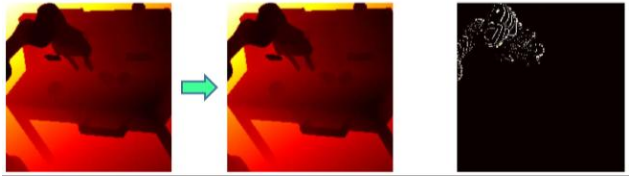

(c) ABC dataset: Extremely wide range

Fig. 5 (a) presents depth map fluctuations at two consecutive time points, illustrating the degree of depth sensitivity through the count of pixel changes. After normalization against the extreme values, the default dataset exhibits 713 pixel variations, whereas the ABC dataset shows 692 pixel variations, indicating a higher depth sensitivity in the default dataset.(b) depicts normalization using a narrow depth range from the default dataset, with bright spots on the right indicating pixel changes within a 0-255 scale. (c) illustrates normalization with a wider depth range from the ABC dataset, resulting in fewer bright spots on a 0-255 scale, suggesting reduced sensitivity to depth variation.

## V. CONCLUSION AND FUTURE WORK

This study introduces an innovative approach to robotic manipulation by effectively integrating depth information with RGB data, significantly enhancing the performance of our model, *RoboFlamingo*. By leveraging depth cues alongside traditional RGB inputs, we have developed a comprehensive model that offers a new paradigm in robotic control. This integration not only breaks the confines of relying solely on RGB data but also enriches the model's environmental perception, leading to improved task success rates. Our findings underscore the pivotal role of depth information in creating more nuanced and capable robotic systems, capable of navigating and interacting with their surroundings more effectively. The success of our approach points towards the exciting potential of utilizing multimodal data for advancing robotic manipulation tasks. Future work will focus on how to transfer simulation data to real-world environments.